\documentclass{article}

\PassOptionsToPackage{sort&compress,numbers}{natbib}

\usepackage[preprint]{neurips_2019}

\usepackage{natbib}
\usepackage[utf8]{inputenc} 
\usepackage[T1]{fontenc}    
\usepackage{hyperref}       
\usepackage{url}            
\usepackage{booktabs}       
\usepackage{amsfonts}       
\usepackage{nicefrac}       
\usepackage{microtype}      
\usepackage[pdftex]{graphicx}
\usepackage{mathtools}
\usepackage{amsmath,amsfonts,amssymb,amsthm}
\usepackage{tabularx}
\usepackage{xspace}
\usepackage[]{appendix}

\DeclarePairedDelimiter\norm{\lVert}{\rVert}%

\DeclareMathOperator*{\argmin}{argmin} 

\title{Towards Interpretable Sparse Graph Representation Learning with Laplacian Pooling}

\author{%
Emmanuel~Noutahi\\
  InVivo AI\\
  \texttt{emmanuel@invivoai.com} 
  \And
  Dominique~Beaini\\
  InVivo AI\\
  \texttt{dominique@invivoai.com}   
  \AND
 Julien~Horwood\\
  InVivo AI, Mila\\
  \texttt{julien@invivoai.com}
  \And 
 Sébastien~Giguère\\
  InVivo AI\\
  \texttt{sebastien@invivoai.com}
  \And 
 Prudencio~Tossou\\
  InVivo AI\\
  \texttt{prudencio@invivoai.com} 
}

\begin{document}

\maketitle

\begin{abstract}

Recent work in graph neural networks (GNNs) has led to improvements in molecular activity and property prediction tasks. Unfortunately, GNNs often fail to capture the relative importance of interactions between molecular substructures, in part due to the absence of efficient intermediate pooling steps. To address these issues, we propose LaPool (Laplacian Pooling), a novel, data-driven, and interpretable hierarchical graph pooling method that takes into account both node features and graph structure to improve molecular representation.
We benchmark LaPool on molecular graph prediction and understanding tasks and show that it outperforms recent GNNs.
Interestingly, LaPool also remains competitive on non-molecular tasks.
Both quantitative and qualitative assessments are done to demonstrate LaPool's improved interpretability and highlight its potential benefits in drug design.
Finally, we demonstrate LaPool's utility for the generation of valid and novel molecules by incorporating it into an adversarial autoencoder.

\end{abstract}

\section{Introduction}

Following the recent rise of deep learning for image and speech processing, there has been great interest in generalizing convolutional neural networks to arbitrary graph-structured data \citep{gilmer_neural_2017, GCN, GIN}. To this end, graph neural networks (GNN) that are either spectral-based or spatial-based approaches have been proposed. Spectral methods define the graph convolution (GC) as a filtering operator of the graph signal \citep{defferrard2016convolutional}, while spatial methods define the GC as a message passing and aggregation across nodes \citep{GCN, GIN, JT_VAE}. For drug discovery applications, GNNs have been very successful across several molecular classification and generation tasks. In particular, they outperform predefined molecular features, known as fingerprints, and string-based approaches for molecular property prediction and the \textit{de novo} generation of compounds having drug-like properties \citep{JT_VAE, MolMP}. 

However, the node feature update performed by GNNs introduces important limitations.
For instance, poor performance of deep GNNs has been observed experimentally and can be attributed to the signal smoothing effect of the GC layers \citep{li_deeper_2018}.
This limits the network's depth and restricts the receptive field of the vertices in the graph to a few-hop neighborhood, which is insufficient to properly capture local structures, relationships between nodes, and the importance of specific subgraphs.
For example, at least three consecutive GC layers are needed to exchange information for atoms at the opposite side of a six carbon atom ring known as benzene.
This issue is further exacerbated by the single global pooling step performed at the end of most GNNs that ignores any hierarchical structure within the graph.
Yet, substructures such as benzenes are ubiquitously found within drugs, natural products are fabricated by living organism from metabolites, synthetic compounds are created by joining chemical fragments by means of chemical reactions, and experienced medicinal chemist describe compounds by their functional groups.
These are all testaments of the importance of having a representation that is suitable for molecular graphs.

To cope with these limitations, graph coarsening (pooling) methods have been proposed to reduce graph size and enable long-distance interaction between nodes. The earliest methods relied on deterministic clustering of the graphs, making them non-differentiable and task-independent \citep{JT_VAE, dafna_learning_2009, tutorial_spectral_clustering, EigenPool}. In contrast, more recent methods use node features but, as we will show, have limited interpretability because they are unable to preserve the graph structures after pooling \citep{DiffPool, graph_UNET}. 

In this work, we focus primarily on molecular graphs which present two particularities: they are often sparse and there is no regularity in the graph signal as adjacent nodes tend to have different features.
Borrowing from theory in graph signal processing, we propose LaPool (Laplacian Pooling), a differentiable pooling method that takes into account both the graph structure and its node features. LaPool uses the graph Laplacian to perform a dynamic and hierarchical segmentation of graphs by selecting a set of centroid nodes as cluster representatives (centroids).
Then, it learns a sparse assignment of the remaining nodes (followers) into these clusters using an attention mechanism. The primary contributions of this paper are summarized below:

\begin{itemize}
\item LaPool is a novel and differentiable pooling module that can be incorporated into existing GNNs to yield more expressive networks for molecular data.
\item LaPool outperforms recently proposed graph pooling layers on broadly used discriminative and generative molecular graph benchmarks, while also performing well on non molecular benchmarks.
\item A first molecular graph structure understanding dataset for benchmarking interpretability of GNNs is proposed.
\item The improved interpretability achieved by LaPool is supported by qualitative and quantitative assessments.
\end{itemize}

We believe that the enhanced performance and interpretability achieved by LaPool will enable improved understanding of molecular structure-activity relationships, having important applications in drug discovery.

\section{Related Work}

In this section, we first introduce related work on graph pooling, then provide an overview of techniques used in computational drug discovery to put our work into context. As our focus herein is on graph pooling, we refer readers to \citet{wu2019comprehensive} for an overview of recent progress in GNNs.

In traditional GNN architectures, global sum, average, and max-pooling layers have been used to aggregate node embeddings into a graph-level representation.
Recently, alternative methods such as gated mechanism, sorting of node features before feeding them into a 1D convolution (SortPool), and Set2Set architecture as replacement for node averaging have been proposed \citep{li2015gated, zhang2018end, gilmer_neural_2017}.
Although these new global aggregation methods have been shown to outperform standard global pooling, they overlook the rich structural information in graphs which has been shown as important for building effective GNN models~\citep{DiffPool,EigenPool}. 

Consequently, hierarchical graph pooling methods have been proposed. They act by reducing graph size and increasing node receptive fields without increasing network depth. However, in contrast to images, graphs have irregular structures, making it challenging to properly pool nodes together. Certain hierarchical graph pooling methods therefore rely on deterministic and non-differentiable clustering to segment the graph \citep{defferrard2016convolutional,JT_VAE}.
More relevant to our work, differentiable hierarchical graph pooling layers have been proposed. DiffPool is as a pooling layer that performs a similarity-based node clustering using a soft affinity matrix learned by a GNN \citep{DiffPool}. Likewise, Graph U-Net was proposed as a sampling method that retains and propagates only the top-k nodes at each pooling step based on a learned importance value \citep{graph_UNET}.

In computer-aided drug discovery, methods such as virtual screening, 
which aims to accurately predict molecular properties directly from molecular structure, can play an important role in rapidly triaging promising compounds early in drug discovery \citep{subramaniam2008virtual}. Importantly, data-driven predictive approaches that leverage advances in deep learning, rather than pre-determined features such as molecular fingerprints \citep{rogers2010extended} and SMILES string representations, have been shown to dramatically improve prediction accuracy \citep{kearnes2016molecular,moleculenet}.

Similarly, advances in generative models have enabled the application of deep generative techniques to propose novel molecules.
These methods are important as the number of feasible molecules, estimated at $10^{60}$, far exceeds those that have been catalogued to date.
The first molecular generative models (e.g. Grammar-VAE \citep{grammarvae}) resorted to generating string representations of molecules via SMILES, which resulted in a high proportion of invalid structures due to the complex syntax of SMILES.
More relevant to our work, graph generative models such as JT-VAE \citep{JT_VAE}, GraphVAE \citep{graphVAE}, MolGAN \citep{molgan}, and MolMP \citep{MolMP} have since been developed and shown to improve the validity and novelty of generated molecules.

\section{Graph Laplacian Pooling}
\begin{figure*}[t]
    \centerline{\includegraphics[width=0.8\linewidth]{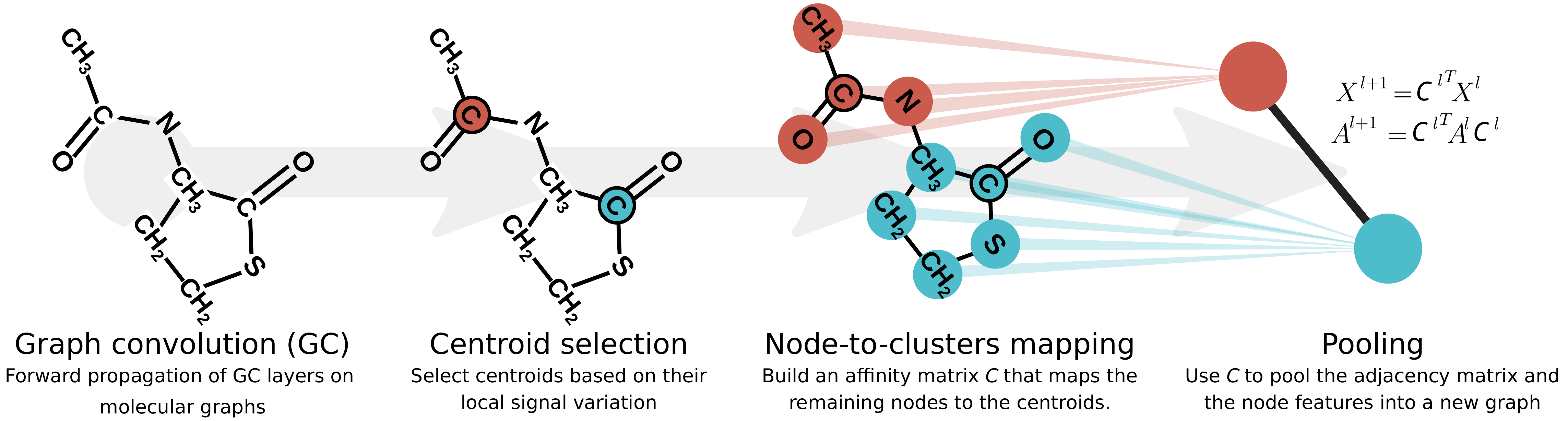}}
  \caption{Overview of the proposed Laplacian Pooling (LaPool) method}
  \label{fig:overview}
\end{figure*}
A reliable pooling operator should maintain the overall structure and connectivity of a graph. LaPool achieves this by taking into account the local structure defined by the neighborhood of each node. As illustrated in Figure~\ref{fig:overview}, the method uses a standard GC layer with a centroid selection and a follower selection step. First, the centroids of the graph are selected based on their local signal variation. 
Next, LaPool learns an affinity matrix $C$ using a distance normalized attention mechanism to assign all nodes of the graph to the centroids. 
Finally, the affinity matrix is used to coarsen the graph. These steps are detailed below.
 
\subsection{Preliminaries}\label{section:prelims}
\textbf{Notation} Let $G = \langle V, A, X \rangle$ be an undirected graph, where $V = \big \{v_1, \dots v_n\big\}$ is its vertex set,  $A \in \{0, 1\}^{n\times n}$ denotes its adjacency matrix, and $X = \big[\bf{x_1}, \dots \bf{x_n}\big]^{\intercal}\, \in \bf{R}^{n\times d}$ is the node feature matrix with each node $v_i$ having $d$-dimensional feature $x_i$.  Note that $X$ can also be viewed as a $d$-dimensional signal on $G$ \citep{shuman2012emerging}. 
Without loss of generality we can assume that $V$, $A$, and $X$ share the same fixed ordering of nodes.
We denote by $\mathcal{N}(v_i)$ the set of nodes adjacent to $v_i$ and by $\mathcal{N}^h(v_i)$ the nodes in the neighborhood of radius $h$ (or $h$-hop) of a node $v_i$, i.e. the set of nodes separated from $v_i$ by a path of length at most $h$.

\textbf{Graph Neural Networks} We consider GNNs that act in the graph spatial domain as message passing \citep{gilmer_neural_2017}. We focus on the Graph Isomorphism Network (GIN) \citep{GIN}, which uses a SUM-aggregator on messages received by each node to achieve a better understanding of the graph structure. 
$\mathbf{x}^l_{i}$ the feature vector for node $v_i$ at layer $l$ is updated using
\begin{equation}\label{eq:GIN}
    \mathbf{x}^l_{i} = \text{M}^l_{\Theta} \left( \mathbf{x}_i^{(l-1)} + \sum_{v_j \in \mathcal{N}(v_i)}{A^{(l-1)}_{i,j} \mathbf{x}_j^{(l-1)}}  \right)
\end{equation} 
where $\text{M}^l_{\Theta}$ is a neural network with trainable parameters $\Theta$ and $v_j \in \mathcal{N}(v_i)$ are the neighbors of $v_i$. Notice that the adjacency matrix $A_{i,j}$ allows one to take into account the edge weight between nodes $v_i$ and $v_j$ when $A$ is not binary.

\textbf{Graph Signal} For any graph $G$, its graph Laplacian matrix $L$ is defined as $L = D - A$, where $D$ is a diagonal matrix with $D_{i,i}$ being the degree of node $v_i$. The graph Laplacian is a difference operator and can be used to define the smoothness $s(X)$ of a signal $X$ on $G$, quantifying the extent at which the signal changes between connected nodes. For 1-dimensional signal $\mathbf{f} = [f_1, \dots, f_n]$ the smoothness is defined as: 
\begin{equation}
s(\mathbf{f}) = ( \mathbf{f}^{\intercal} L \mathbf{f}) = \frac{1}{2}\sum\limits_{i,j}^n  A_{i,j}( f_i - f_j)^2
\end{equation}

\subsection{Graph Downsampling}
\label{section:leaders}

This section details how LaPool downsamples the original graph by selecting a set $V_C$ of nodes as centroids after $l$ consecutive GC layers.

\textbf{Centroid Selection}
For each vertex $v_i \in V$, we define $s_i$ as its local measure of intensity of signal variation around $v_i$. Intuitively, $s_i$ measures how different the signal residing at a node $v_i$ is from its neighbors. Formally, we use a definition of local signal variation similar to the \textit{local normalized neighboring signal variation} described in \citep{chen2015signal}, with the only difference being the absence of degree normalization:
\begin{align}\label{eq:leaders}
s_i &= \big\| \smashoperator{\sum_{j\in \mathcal{N}(v_i)}}A_{i,j}(\mathbf{x}_i - \mathbf{x}_j)\big\|_2 & \nonumber \\  S &=  \big[ s_1, \dots s_n \big]^{\intercal} = \| LX \|_{2,\mathbf{R}^d}
\end{align}
where $||\cdot||_{2, \mathbf{R}^d}$ corresponds to taking the vector norms over the d-dimensional rows of $LX$.
As we are interested in the set of nodes that have an intensity of signal variation greater than their neighborhood, we define $V_C = \text{top}_V(LS, k)$ as corresponding to the top $k$ nodes having greatest intensity of signal variation.
Similarly, signal variation for the $h-$hop neighborhood of a vertex can be computed by taking the power of the Laplacian in Eq. \ref{eq:leaders}.

Observe that the GC layers preceding each pooling step perform a smoothing of the graph signal and thus act as a low-pass filter. Instead, Eq. (\ref{eq:leaders}) emphasizes the high variation regions, resulting in a filtering of $X$ that attenuates low and high-frequency noise, yet retains the important signal information. The intuition for using the Laplacian maxima for selecting the centroids is that a smooth signal can be approximated well using a linear interpolation between its local maxima and minima. This is in contrast with most approaches in Graph Signal Processing that use the lower frequencies for signal conservation but requires the signal to be k-bandlimited \citep{EigenPool, chen_signal_2015, chen_discrete_2015}. For a 1D signal, LaPool selects points, usually near the maxima/minima, where the derivative changes the most and is hardest to interpolate linearly. For molecular graphs, this often corresponds to sampling a subset of nodes critical for reconstructing the original molecule.


\textbf{Dynamic Selection of the Centroids} The method presented in the previous section implies the selection of $k$ centroids. Because the optimal value for $k$ can be graph-dependent and an inappropriate value can result in densely located centroids, we propose an alternative method to dynamically select the nodes with signal variation $s_i$ greater than its neighbors $s_j$:
\begin{equation}\label{eq:dyn_leaders}
 V_C =  \{v_i \in V \mid \forall\; v_j,  s_i - A_{ij} s_j  > 0  \}  
 \end{equation}
This offers the unique flexibility of determining the clustering dynamically, when training graphs sequentially, or statically when performing batch training.

\subsection{Learning the Node-to-Cluster Assignment Matrix}
\label{section:followers}

Once $V_C$, the set of centroid nodes, is determined, we compute a mapping of the remaining ``follower'' nodes $V_F = V \setminus V_C$ into the clusters formed by the nodes in $V_C$.  This mapping gives the cluster assignment 
$C = \big[ \mathbf{c}_1, ... \mathbf{c}_n \big]^{\intercal}\in [0,1]^{n\times m}$ s.t. $\forall i : \mathbf{1}\mathbf{c}_i^\intercal = 1$, where each row $\mathbf{c}_i$
corresponds to the affinity of node $v_i$ towards each of the $m$ centroids in $V_C$.  

Let $X$ be the node embedding matrix and $X_{C}$ the embedding of the ``centroids'' at an arbitrary layer.  We compute  $C$ using a soft-attention mechanism \citep{graves2014neural} measured by the cosine similarity between $X$ and $X_C$ :
\begin{equation}\label{eq:attention}
    \mathbf{c}_i = \begin{cases}
    \delta_{i,j} &   \text{if  $v_i \in V_C$} \\
     \text{sparsemax} \Big(\beta_i \frac{\mathbf{x_i} \cdot X_C}{\norm{ {\mathbf{x_i}} }   \norm{X_C}} \Big) & \text{otherwise}
    \end{cases}
\end{equation} 
where $\delta_{i,j}$ is the Kronecker delta and prevents the selected centroid nodes from being assigned to other clusters. The sparsemax operator \citep{laha_softmax_2016, martins_softmax_2016} is an alternative to the softmax  operator and is defined by
$\text{sparsemax}(\mathbf{z}) = \argmin\limits_{\mathbf{p} \in \Delta^{K-1}} \| \mathbf{p}- \mathbf{z}\|^2$. It corresponds to the euclidean projection of $\mathbf{z}$ onto the probability simplex  $\Delta^{K-1} = \big\{\mathbf{p} \in \mathbb{R}^{K}\,|\,\mathbf{1}^\intercal\mathbf{p} = 1, \mathbf{p} \geq 0 \big\}$. The sparsemax operator ensures the sparsity of the attention coefficients, encourages the assignment of each node to a single centroid, and alleviates the need for entropy minimization as done in DiffPool. 
The term $\beta_i$ allows us to regularize the value of the attention for each node. We can define $\beta_i = \frac{1}{\bf{d}_{i,V_C}}$, where $\bf{d}_{i,V_C}$ is the shortest path distance between each node $v_i \in V_F$ and centroids in $V_C$.  Although this regularization incurs a cost $\mathcal{O} ({|V|^2} |V_C|)$, it strengthens the affinity to closer centroids and ensures the connectivity of the resulting pooled graph. This regularization is used as a hyperparameter of the layer and can be turned off or, alternatively, the mapping can be restricted to centroids within a fixed $h-$hop neighborhood of each node. 

Finally, after $C^l$ is computed for a given layer $l$, the coarsened graph $G^{(l+1)} =   \langle V^{(l+1)},  A^{(l+1)}, X^{(l+1)}\rangle$ is computed, as in \citep{DiffPool}, using
\begin{align} \label{eq:projection}
  A^{(l + 1)} &= {C^{(l)}}^{\intercal} A^{(l)}  C^{(l)} \in \mathbb{R}^{|V^{(l)}_C|\times|V^{(l)}_C|} \nonumber \\
  X^{(l + 1)} &= {M_\Psi}\big({C^{(l)}}^\intercal X^{(l)}\big)
\end{align}
where $M_\Psi$ is a neural network with trainable parameters $\Psi$ and is used to update the embedding of nodes in $G^{(l+1)}$ after the mapping. This process can be repeated by feeding the new graph $G^{(l+1)}$ into another GNN layer.

In summary, this allows for a node-to-cluster assignment approach that retains the feature content of the other nodes in a graph via the soft-assignment of followers to their centroids. As we will show in the Results section, the node clusters can be used to identify graph substructures that are important with respect to a prediction task.

\section{Results and Discussion}

Our primary objective while developing LaPool was to learn an interpretable representation of sparse graphs, notably molecular substructures. We argue that this is an essential step towards shedding light into the decision process within neural networks and ultimately increasing their utility in the design of new drugs. This implies that GNNs should be able to identify semantically important substructure components from molecular graphs, and eventually reconstruct these graphs from such components. This stems from the intuition that molecular validity and functional properties derive more from chemical fragments than individual atoms.

Our experiments therefore aim to empirically demonstrate the following properties of LaPool, as benchmarked against current state-of-the-art pooling models and the Graph Isomorphism Network:

\begin{itemize}
    \item LaPool's consideration of semantically important information such as node distance translates to improved performance on molecular understanding and molecular activity prediction tasks.
    \item Visualization of LaPool's behaviour at the pooling layer demonstrates its ability to identify coherent and meaningful molecular substructures.
    \item The hierarchical representation enforced by LaPool, which preserves the original graph structure, improves model interpretability.
    \item Learning meaningful substructures can be leveraged to construct a generative model which leads to more realistic and feasible molecules.
\end{itemize}

Throughout our experiments, we use the same architecture for all models to ensure a fair comparison across all pooling layers: 2 layer GNNs with 128 filters each before the optional pooling layer, followed by 2 GNNs with 64 filters each and two fully connected layers including the output layer. Detailed information on architectural tuning and pooling-specific hyper-parameter search are provided in Supplemental Section~A.

\subsection{Benchmark on Molecular Graph Understanding}\label{section:subpred}

DiffPool and Graph U-Net models have been shown to outperform standard graph convolution networks
on several graph benchmark datasets \citep{DiffPool, graph_UNET}. Although not explicitly stated, both methods are most effective when the graph signal is smooth. In cases where adjacent nodes tend to be similar, the DiffPool procedure will cluster together nodes in the same neighborhood, maintaining the overall graph structure, while Graph U-net will select nodes in the same neighborhood and will not create isolated components that no longer exchange information. However, on molecular graphs, the graph signal is rarely smooth. Therefore, we expect these two methods to be less effective at identifying molecular substructures, given that they do not explicitly consider structural relationships.

We demonstrate this empirically by evaluating these methods at identifying known molecular substructure information from publicly available molecular datasets.
We use a subset of approximately 17,000 molecules extracted from the ChEMBL database \citep{li2018multi} and benchmark all methods at predicting the presence of 86 molecular fragments arising purely from structural information, as well as 55 structural alerts associated with molecule toxicity. These new datasets are provided alongside our source code.

As shown in Table~\ref{tab:fragments}, LaPool globally outperforms other pooling models and GIN for the F1 and ROC-AUC metrics (micro-averaged to account for high class imbalance). In particular, on the harder and extremely imbalanced molecular alerts prediction task, all models performed poorly compared to LaPool, suggesting that the hierarchical representation learned by LaPool helps to achieve a better understanding of the molecular graphs.
\begin{table*}[ht]
    \centering
    \caption{Molecular fragment and structural alert prediction results on ChEMBL dataset (5-fold cross-validation).} \label{tab:fragments}
    \begin{tabular}{lcccc}
\toprule
{} & \multicolumn{2}{c}{Fragments} & \multicolumn{2}{c}{Alerts} \\\cmidrule{2-5}
{} &      F1 &   ROC-AUC &         F1 &   ROC-AUC \\
\midrule
GIN          &  \textbf{99.436 $\pm$ 0.545} &   \textbf{99.991  $\pm$  0.004} &   31.759 $\pm$ 3.728 &      82.495 $\pm$ 8.429 \\
DiffPool     & 97.961 $\pm$ 0.384	 & 99.967 $\pm$ 0.025 & 48.638	 $\pm$ 9.916 &   76.537    $\pm$ 0.241 \\
Graph U-net &   95.469 $\pm$  1.414 &  99.962 $\pm$ 0.033	 &  37.585    $\pm$ 2.978 &  85.124   $\pm$ 8.447 \\
LaPool & 98.980  $\pm$  0.506 & \textbf{99.994	$\pm$ 0.000}	 & \textbf{78.592 $\pm$ 7.217} &  \textbf{94.164 $\pm$ 1.784}	 \\
\bottomrule
\end{tabular}
\end{table*}{}

\subsection{Experiments on Standard Graph Classification Benchmarks}
In addition, we benchmark our model on molecular toxicity prediction using the well known TOX21 dataset \citep{national2007toxicity}.
To assess the performance of LaPool outside the domain of molecular graphs, we conducted experiments on datasets containing larger or denser graphs: DD, PROTEINS, and FRANKENSTEIN.
Datasets statistics are given in Supplemental Section C.
For TOX21, we report the test ROC-AUC averaged over 5-folds, following the 80-10-10 split proportion used in ~\citet{moleculenet}, while we follow prior work \citep{DiffPool} on the remaining datasets by reporting the best accuracy on a 10-fold cross-validation.

As shown in Table \ref{tab:supervised}, LaPool compares favourably with other approaches on the TOX21 molecular dataset and on the PROTEINS and FRANKENSTEIN, both of which contain non-molecular graphs with size similar to the TOX21 molecules. In particular, on the PROTEINS dataset LaPool achieved an accuracy of $83.83$, representing a significant gap relative to DiffPool, its closest competitor, at $77.25$.
This suggests that the LaPool method is not restricted to molecular data and is expected to have broad applicability, especially in the context of sparse graph classification.

\begin{table*}[htp]
\centering
\caption{Performance evaluation on standard benchmark graphs}
\label{tab:supervised}
\small\begin{tabular}{lc|ccc}
\toprule
{} & {TOX21} &  DD &   PROTEINS &  FRANKENSTEIN \\
\midrule
GIN          &  82.90
 $\pm$ 0.69
 \   &  77.97 &  66.47 &     68.20 \\
DiffPool   & 82.37
 $\pm$ 0.90 & \textbf{85.88} &  77.25 &     69.12 \\
Graph U-net &  81.41
 $\pm$ 0.60  &  77.40 &  74.50 &     66.16\\
LaPool  & \textbf{83.42 $\pm$ 0.97}  &  81.36 &  \textbf{83.83} &   \textbf{ 69.74 }\\
\bottomrule
\end{tabular}
\end{table*}

\subsection{Improved Interpretability}

To better understand the insights provided by LaPool, we conduct model interpretability experiments on molecular fragment prediction. Here we refer to interpretability as the degree to which a human (in this context, a medicinal chemist) can understand the cause of the model’s decision \citep{miller2018explanation}. This justifies our focus on fragment prediction since, in that setting, an interpretable model that achieves high performance would need to first understand the graph structure and the relationship between nodes.

\begin{figure*}[hbt]
    \centerline{\includegraphics[width=1\textwidth]{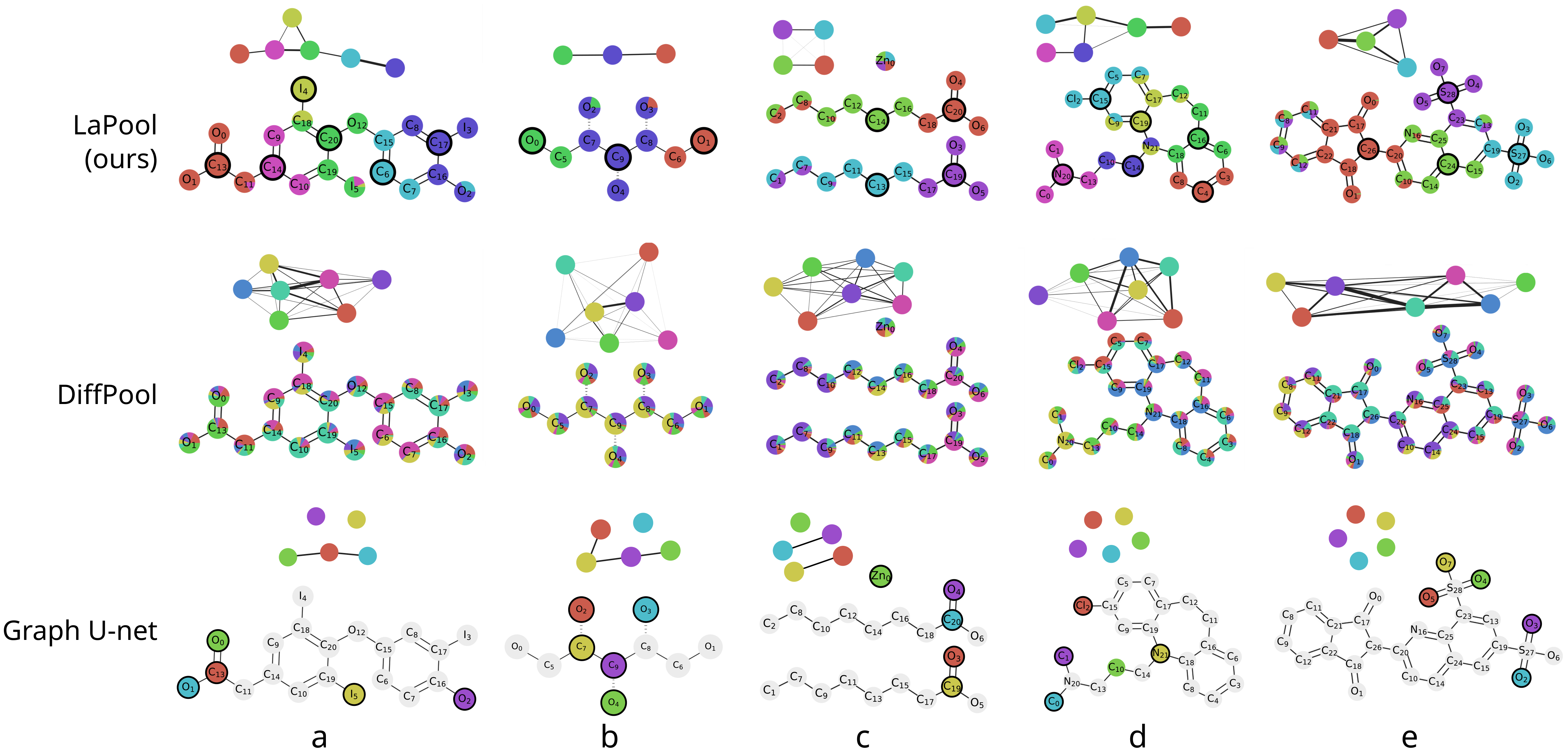}}
    \caption{Visualization of node selection and/or clustering performed by DiffPool, Graph U-net and LaPool for structural alert prediction. Dynamic clustering was used for LaPool, while the cluster size ($k$) resulting in the the best model was used for DiffPool($k=7$) and Graph U-net ($k=5$). The top graph is the pooled graph. The bottom graph is the original molecule, with the pie-charts representing the cluster affinity of each node. For LaPool and Graph U-net, the bold nodes represent the chosen centroids and selected nodes, respectively.}
    \label{fig:interpretability}
\end{figure*}

We first investigate the behavior of LaPool, DiffPool and Graph U-net by analyzing the clustering, or graph downsampling, made at the pooling layer level. We argue that an interpretable pooling layer should preserve the overall graph structure after pooling and produce \textit{meaningful} clusters that could provide insight into the contribution of each molecular subgraph. While defining what is meaningful is inherently subjective, we attempt to shed light on these models by observing their behavior in the drug discovery domain, using well understood chemical structures as reference.

We show in Figure \ref{fig:interpretability} that LaPool is able to coarsen the molecular graphs into sparsely connected graphs, which can be interpreted as the \textit{skeleton} of the molecules. Indeed, the data-driven dynamic segmentation it performed is akin to chemical fragmentation \citep{gordon2011fragmentation}.
In contrast, DiffPool's cluster assignment is more uniform across the graph, leading to densely connected coarsened graphs which are less interpretable from a chemistry viewpoint. In particular, it fails in the presence of molecular symmetry, as it encourages the mapping of nodes with similar features to the same clusters. This is illustrated in example (c) which shows how DiffPool creates a fully connected graph from an originally disconnected graph, and example (b) which shows how symmetric elements, despite being far from each other, are assigned identically. On the other hand, we observe that Graph U-net ignores the graph structure, typically disconnecting it. It also appears biased toward selecting atoms in a similar environment to ones already selected. Such failures are not present when using LaPool, since the dynamic centroid selection and the subsequent distance regularization enforce preservation of the molecular graph structure. A failure case for LaPool is seen in (e) and corresponds to a missing centroid node in the left region of the graph, which results in a soft assignment of the region to multiple clusters.

\begin{figure*}[htb]
\centering
\includegraphics[width=0.9\linewidth]{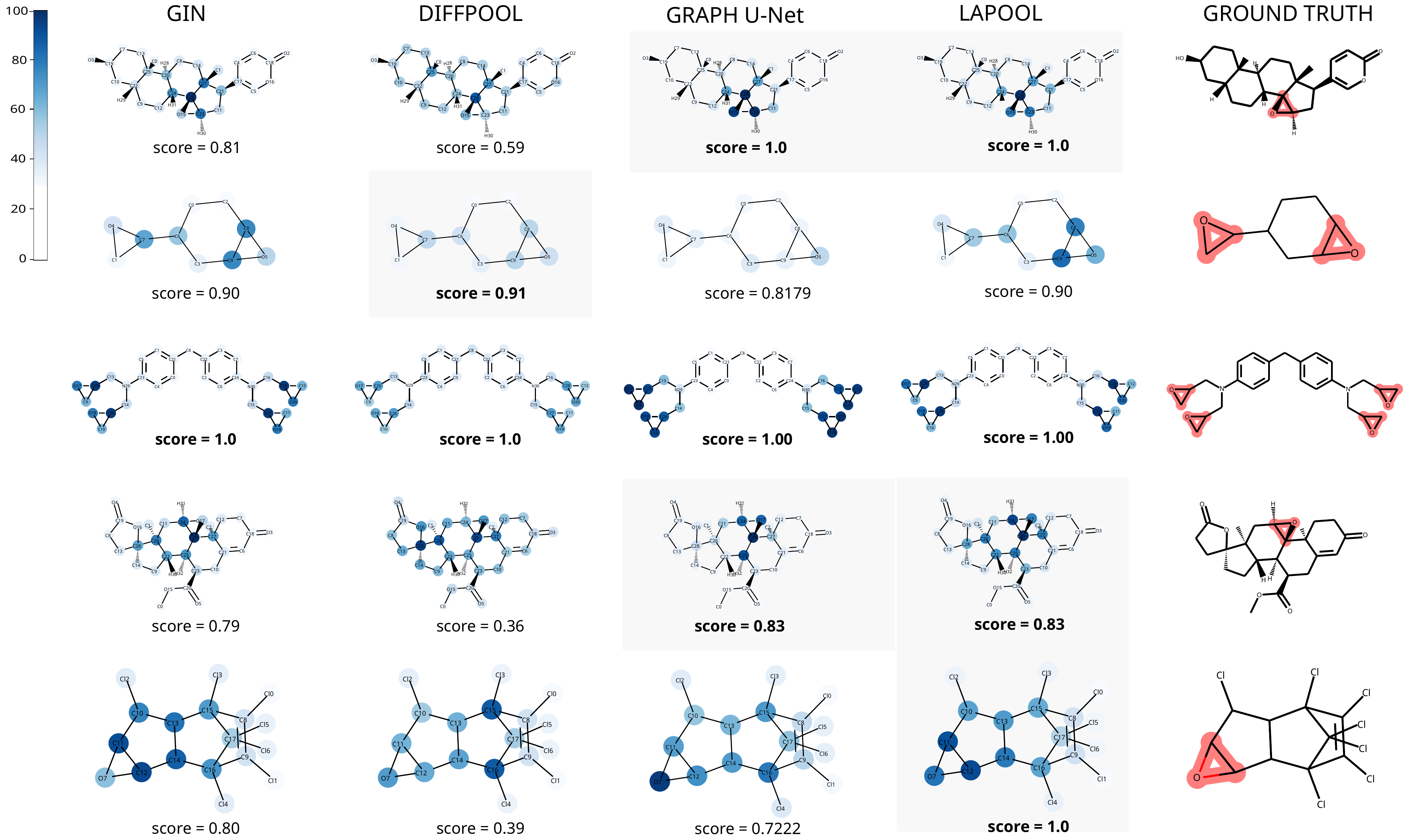}
  \caption{Example of node importance for  GIN, DiffPool, Graph U-Net and LaPool models. The ground truth is shown on the right and nodes are highlighted according to their predicted importance. The computed interpretability score is indicated below each molecules and the best performing models on each example are highlighted in gray. The computed interpretability score is shown for each model.}\label{fig:explain}
\end{figure*}

We also attempt to quantify interpretability by computing an explanation of each model decision and comparing it to a ground truth. We design a simple experiment in which we predict the presence of either Epoxide, Thioepoxide, or Aziridine substructures that are indicative of molecular toxicity and are denoted by the molecular pattern ``C1[O,S,N]C1''. As a surrogate for interpretability, we compute the accuracy of the importance attributed by each model to relevant substructures of the input molecules, given the presence of the fragment we wish to predict. Similar to \citet{pope2019explainability}, we adapt an existing CNNs explainability method to GNNs. Specifically, we choose to compute the integrated gradient \citep{DBLP:journals/corr/SundararajanTY17} over the input node features due to its stability and robustness in the presence of zero-value features (see Supplemental Section~C for discussion and alternate approaches). Next, we derive an importance score for each node using the L1-norm of the feature attribution map for the node.

To assess the ability of each method to distinguish between important and non-important nodes, Table~\ref{tab:explain} presents PR-AUC values computed using the node importance values and the ground truth node assignment. We find that LaPool achieves the highest PR-AUC values, suggesting that it identifies the salient structure more robustly. An interesting outcome of this experiment is the performance of Graph U-net which, despite its decimation of the graph, ranked second best. This is likely a direct result of using a cluster size large enough to cover the toxic fragment size. Indeed, given that the importance values are computed at the node input level, Graph U-net should perform well if its projection vector selects the relevant fragment atoms.  
Finally, we provide visual examples of node importance for all methods in Figure~\ref{fig:explain}.

\begin{table}[h]
\centering
\caption{Comparison of prediction explainability based on average PR-AUC over node attribution}
\label{tab:explain}
\footnotesize\begin{tabular}{lcccc}
\toprule
 & GIN & DiffPool & Graph U-Net & LaPool  \\
\midrule
 avg. PR-AUC &   0.876 & 0.799 & 0.879 & \textbf{0.906} \\
\bottomrule
\end{tabular}
\end{table}

\subsection{Molecular Generation}
We aim to showcase LaPool's utility in drug discovery by demonstrating that it can be leveraged to generate molecules. In previous work, GANs and VAEs were used to generate either string or graph representations of molecules. Here, we use the GAN-based Wasserstein Auto-Encoder \citep{wae} to model the data distribution of molecules (see Figure~1 in Supplemental Material). For the encoder, we use a similar network architecture as in our supervised experiments. The decoder and discriminator are simple MLPs, with complete architectural details provided in Supplemental Section A.4. Although the encoder is permutation invariant, the decoding process may not be. To force the decoder to learn a single node ordering, we use a canonicalization algorithm that reorders atoms to ensure a unique graph for each molecule \citep{schneider2015get}. 
We further improve the robustness of our generative model to node permutations by computing the reconstruction loss using a permutation-invariant embedding, parameterized by a GIN, on both the input and reconstructed graphs (see Supplemental Section A.4.2). We find that such a formulation improves the reconstruction loss and increases the ratio of valid molecules generated.

\textbf{Dataset and Baseline Models} Following previous work on molecular generation, we evaluate our generative model with an encoder having a LaPool layer (referred to as WAE-LaP) on the QM9 molecular dataset \citep{ramakrishnan2014quantum}. This dataset contains 133,885 small drug-like organic compounds with up to 9 heavy atoms (C, O, N, F). We compare WAE-LaP to alternatives within our WAE framework where either no pooling is used (WAE-GNN) or where DiffPool is used as the pooling layer (WAE-Diff). Our results are also compared to previous results on the same dataset, including Grammar-VAE, GraphVAE, and MolGAN.

\textbf{Evaluation Metrics} We measure the performance of the generative models using metrics standard in the field: proportion of valid molecules from generated samples (validity), proportion of unique molecules generated (uniqueness), and proportion of generated samples not found in the training set (novelty). All metrics were computed on a set of 10,000 generated molecules.
\begin{table*}[htb]
\caption{Performance comparison of the generative models on QM9. Values are reported in percentages and baseline results are taken from~\citep{molgan}.}
  \label{table:gen}
  \centering
    \begin{tabular}{lccc|ccc}
    \toprule
 & WAE-GNN & WAE-Diff & WAE-LaP & Grammar-VAE & GraphVAE & MolGAN \\ \midrule
\% Valid  & 96.8 & 97.2 & \textbf{98.8}  & 60.2 & 91.0 & 98.1 \\
\% Unique & 50.0 & 29.3 & \textbf{65.5} & 9.3  & 24.1 & 10.4 \\
\% Novel  & 78.9 & 78.9 & 78.4 & 80.9 & 61.0 & \textbf{94.2}  \\
\bottomrule
\end{tabular}
\end{table*}

Empirical results are summarized in Table~\ref{table:gen}.
Regarding validity of the generated molecules, WAE-LaP performed best compared to both the WAE-based and previously proposed generative approaches. We hypothesize that this can be attributed to LaPool allowing the learning of a more robust graph representation, in turn making the decoding task easier.
Regarding uniqueness of the molecules, WAE-LaP significantly outperformed other WAE-based methods by an increase of $+15.5 \%$ and $+36.2 \%$, clearly demonstrating the benefit of the LaPool layer on that metric. WAE-LaP also outperformed previously proposed methods including MolGAN, a method that performs well on other metrics, by $+55.1 \%$.
Regarding novelty of the molecules, WAE-LaP performed similarly ($-0.5 \%$) to other WAE-based methods but was outperformed by MolGAN by $+15.8 \%$.
We hypothesize that the decrease in novelty observed with LaPool might be a result of its pooling mechanism encouraging fragment novelty during sampling, thus limiting novelty resulting from rearrangement at the atom level.
Alternatively, the difference in novelty could be explained by the choice of generative model but exploring alternatives to the WAE was not in the scope of this paper as our objective was not to propose a new molecular generative method.
Nevertheless, we conclude that the pooling performed by LaPool can improve molecular graph representation, which is crucial in a generative setting. 

\section{Conclusion}
In this work, we have proposed LaPool, a differentiable and robust pooling operator for molecular and sparse graphs that considers both node information and graph structure. In doing so, we have proposed a method capable of identifying important graph substructures by leveraging the graph Laplacian. In contrast with previous work, this method retains the connectivity structure and feature information of the graph during the coarsening procedure, while encouraging nodes belonging to the same substructure to be mapped together in the coarsened graph. 

By incorporating the proposed pooling layer into existing graph neural networks, we have demonstrated that the enforced hierarchization allows for the capture of a richer and more relevant set of features at the graph-level representation. We discussed the performance of LaPool relative to existing graph pooling layers and demonstrated on molecular graph classification and generation benchmarks that LaPool outperforms existing graph pooling modules and produces more interpretable results. In particular, we argue that the molecular graph segmentation performed by LaPool provides greater insight into the molecular cause and that the associated properties can be leveraged in drug discovery. Finally, we show that although LaPool was designed for molecular graphs, it generalizes well to other graph types. In future work, we aim to investigate how additional sources of information such as edge features could be incorporated into the graph pooling process.

\small
\bibliography{biblio.bib}
\newpage
\begin{appendices}
\renewcommand\thefigure{\thesection.\arabic{figure}}    
\section{Network Architecture and training procedure}\label{archsearch}

Below, we describe the network architecture and the training procedure used for both supervised and generative experiments.

\subsection{Edge attributes}\label{edgelayer}
Part of the work presented assumes the absence of edge attributes in the graphs. However, in molecular graphs, the nature of a bond between two atoms plays an important role regarding activity and property. As such, edge types should be considered, especially in generative models. To consider this, we add to our network an initial Edge-GC layer described in the following. 

Let $G = \langle V, E, X \rangle$ be an undirected molecular graph, such that $E = [E_1, \dots E_k] \in \{0,1\}^{e \times n\times n}$ where $n$ is the number of nodes in the graph and $e$ is the number of possible edge.
We have that
\begin{equation}
\sum\limits_{1\leq i\leq e} E_{::i}  = A \in \{0,1\}^{n\times n}
\end{equation}
where $A$ is the adjacency matrix of the graph. The Edge GC layer is simply defined as : 

\begin{equation}
    Y = M_{\Theta_1} (E_1, X) \Vert \dots \Vert M_{\Theta_e} (E_e, X)
\end{equation}

where $\Vert$ is the concatenation operator on the node feature dimension and $M_{\Theta_1}$ are graph neural networks parameterized to learn different features for each edge type. A new graph defined as $G' = \langle V, A, Y \rangle$ can then be feed into the subsequent layers of the network.

\subsection{Atom and edge features}

In our experiments, the initial node feature tensor is represented by a one-hot encoding of atoms (ignoring hydrogens) within the respective datasets and additional properties such as the atom implicit valence, its formal charge, number of radical electrons and whether it is in a molecular ring.  For edge attributes, we consider the single, double and triple bond, which were enough to cover all molecules in the datasets, given that feature extraction was preceded by kekulization of molecules.

\subsection{Supervised experiments}

In all of our supervised experiments, we use a graph convolution module consisting of two graph convolutional layers of 128 channels each with ReLU activation; followed by an optional hierarchical graph pooling layer; then two additional  graph convolution layers (64) with skip connection to introduce jumping knowledge and a gated global graph pooling layer \citep{li2015gated} to yield a graph-level representation. This is further followed by one fully connected layers (128) with batch normalization and ReLU activation, finalized by a linear output layer with appropriate activation for the task readouts. Notice that we used one pooling layer, since no noticeable improvement was observed when using more in our experimental setting.

For DiffPool, we performed a hyperparameter search to find the optimal number of clusters ({12.5\%, 25\%, 50\%} of the maximum number of nodes in the batch~\citep{DiffPool}). A similar search is also performed for the Graph U-net  pooling layer. For LaPool, we consider the same number of clusters and the dynamic node seelction. We also performed a grid search over the window size $k$ used as regularization to prevent nodes from mapping to centroids that are more than $k$-hop away as an alternative to the distance-regularized version.
The grid search was performed for $k \in \{ 1, 2, 3\}$.

For the supervised experiments, we use a batch size of 64 and train the networks for 100 epochs, with early stopping.

\begin{figure*}[htb]
    \centering
    \includegraphics[width=0.9\linewidth]{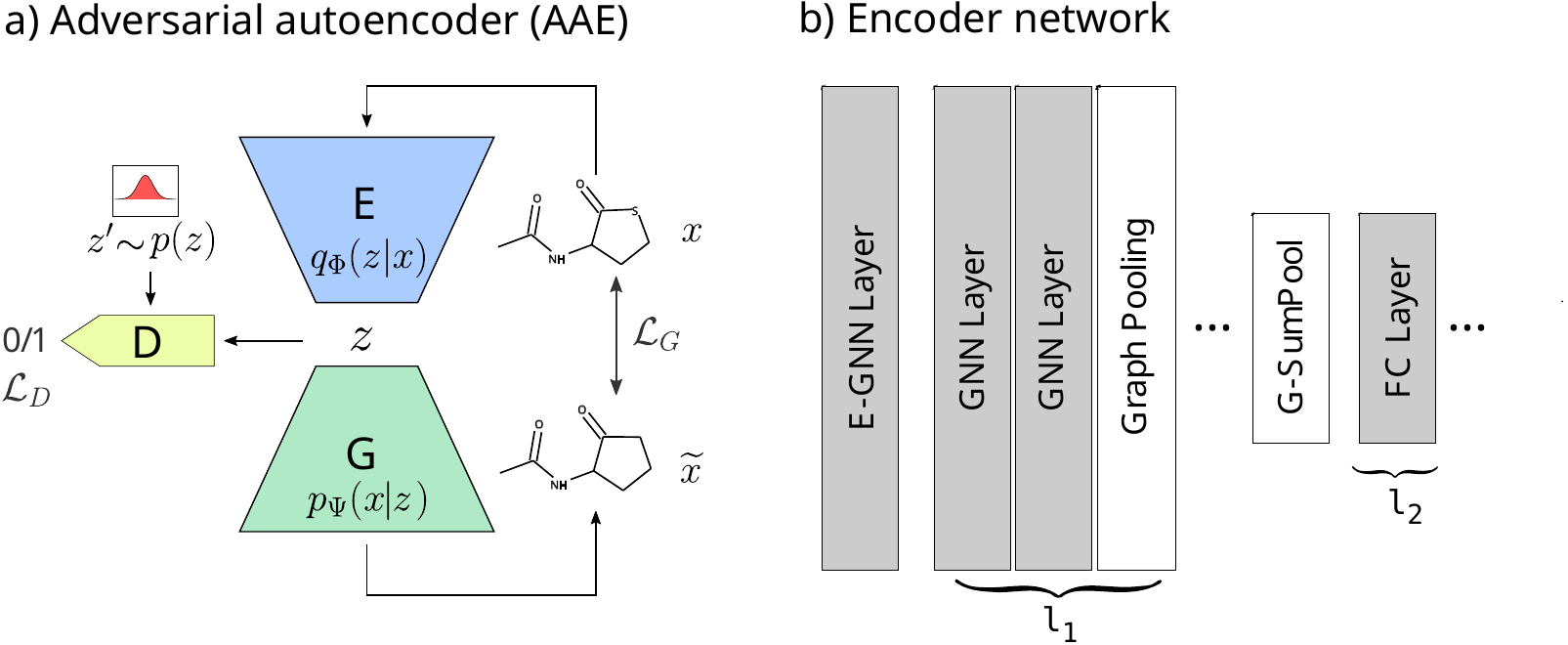}
  \caption{Model architecture for the generative model. (a) We use a WAE, in which a generator (auto-encoder) progressively learns the true molecular data distribution. (b) Architecture used for the encoder network.}
  \label{fig:architecture}
\end{figure*}

\subsection{Generative models}\label{waemodel}

\subsubsection{WAE model}

We use a Wasserstein Auto-Encoder (WAE) as our generative model (see Figure~\ref{fig:architecture}). The WAE minimizes a penalized form of the Wasserstein distance between a model distribution and a target distribution, and has been shown to improve learning stability. As described in \citet{wae}, we aim to minimize the following objective:
\begin{equation}
\inf\limits_{Q(Z|X) \in \mathcal{Q}} \mathbb{E}_{P_X} \mathbb{E}_{Q(Z|X)} [cost (X, G(Z)] + \lambda \mathcal{D}_Z(Q_Z, P_Z) 
\end{equation}
where $\mathcal{Q}$ is any nonparametric set of probabilistic encoders, $D_Z$ is the Jensen-Shannon divergence between the learned latent distribution $Q_Z$ and prior $P_Z$, and $\lambda > 0$ is a hyperparameter. $D_Z$  is estimated using adversarial training (discriminator).

For our generative model, the encoder follows a similar structure as the network used for our supervised experiments, with the exception being that the network now learns a continuous latent space $q_\Psi (z|\mathcal{G})$ given a set of input molecular graphs $\mathcal{G} = \{G_1, \cdots, G_n\}$. More precisely, it consists of one edge graph layer, followed by two GCs (32 channels each), an optional hierarchical graph pooling, two additional GC layers (64, 64), then one global sum pooling step (128) and two fully connected layers (128), meaning the molecular graphs are embedded into a latent space of dimension 128. 
Instead of modeling the node/edge decoding with an autoregressive framework as done in recent works \citep{graphrnn,defactor,Li2018a} to capture the interdependency between them, we used a simple MLP that takes the latent code $z$ as input an pass it through two fully connected layers (128, 64). The output of those layers is used as shared embedding for two networks: one predicting the upper triangular entries of the edge tensor, and the second predicting the node features tensor. This results in faster convergence.

For the discriminator, we use a simple MLP that predicts whether the latent code comes from a normal prior distribution $z ~  \mathcal{N}(0, 1)$. This MLP is constituted by two stacked FCLs (64, 32) followed by an output layer with sigmoid activation. As in \citet{kadurin2017drugan}, we do not use batch-normalization, since it resulted in a mismatch between the discriminator and the generator.

All models use the same basic generative architecture, with the only difference being the presence of a pooling-layer and its associated parameters. For DiffPool, we fixed the number of cluster to three, while for LaPool, we use the distance-based regularization.

\subsubsection{Reconstruction loss}\label{recloss}

For each input molecular graph $G = \langle V, E, X \rangle$, the decoder reconstruct a graph $\tilde{G} = \langle \tilde{V}, \tilde{E}, \tilde{X} \rangle$.
Since we use a canonical ordering (available in RDKit) to construct $G$ from the SMILES representation of molecules, the decoder is forced to learn how to generate a graph under this order. Therefore, the decoding process is not necessarily able to consider permutations on the vertices set, and generation of isomorphic graphs will be heavily penalized in the reconstruction loss. In \citet{graphVAE}, the authors use an expensive graph matching procedure to overcome that limitation. We argue that it suffices to compute the reconstruction loss on $\gamma(G)$ and $\gamma(\tilde{G})$, where $\gamma$ is a permutation invariant embedding function. As a heuristic, we used a Graph Isomorphism Network (GIN), with weights fixed to 1, in order to approximate the Weisfeiler-Lehman graph isomorphism test (see \citet{GIN} for more details). In particular, we use an edge-aware GIN layer (see section~\ref{edgelayer}) to embed both $G$ and $\tilde{G}$. The reconstruction loss is then defined as:

\begin{equation}
\mathcal{L}_{rec} = \frac{1}{|V|} \sum_i (\gamma(G)_i - \gamma(\tilde{G})_i)^2
\end{equation}

Our experiments show that this loss function was able to produce a higher number of valid molecules, although we speculate that such a heuristic might prove harder to optimize on datasets with larger graphs.

\subsubsection{Training procedure}
The QM9 dataset was split into a train (60\%), valid (20\%) and a hold-out test dataset (20\%).
Note that only 25\% of the training set is sampled during each epoch (batch size 32). The generator network (encoder-decoder) and the discriminator network are trained independently, using the Adam optimizer~\cite{kingma2014adam} with an initial learning rate of $1e-4$ for the generator and $1e-3$ for the discriminator. During training, we slowly reduce the learning rate by a factor of 0.5, for the generator, on plateau. To stabilize the learning process and prevent the discriminator from becoming "too good" at distinguishing the true data distribution from the prior, we train the generator two times more often.

\section{Generated molecules}\label{molgenres}

\begin{figure*}[htb]
    \centering
    \includegraphics[width=1\linewidth]{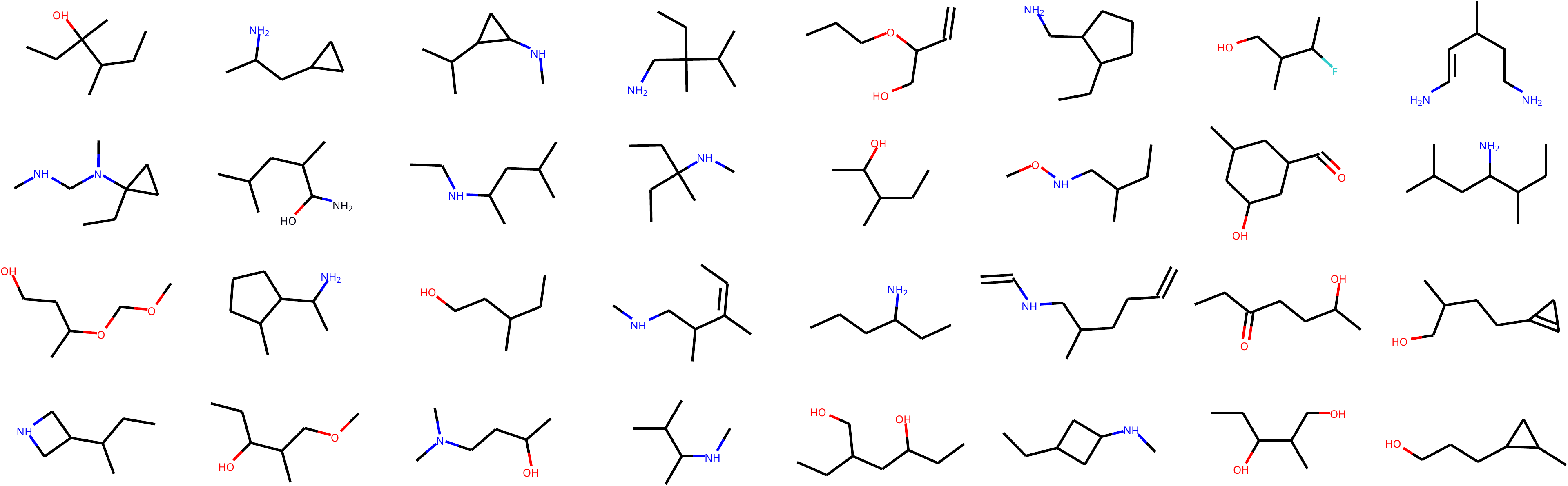}
  \caption{Example of molecules generated by WAE-LaP. Hydrogen atoms are not shown for simplicity.}
  \label{fig:molgen}
\end{figure*}

Figure \ref{fig:molgen} highlights a few molecules generated by WAE-LaP on the QM9 dataset.

\section{Dataset statistics}

Table \ref{table:dataset_statistics} shows dataset statistics and properties.

\begin{table}[htb]
\centering
\caption{Dataset statistics and properties.}
\label{table:dataset}
\footnotesize\begin{tabular}{lcccc}
\toprule
 & TOX21 & DD & PROT. & FRANKEN.   \\
\midrule
 Avg. nodes &   18.51 & 284.32 & 39.05 & 16.83 \\
Avg. edges &   19.23 & 715.66 & 72.82 & 17.88 \\

\#Graphs &   8014 & 1178 & 1113 & 4337 \\
\#Classes &   12 & 2 & 2 & 2\\
\bottomrule
\end{tabular}
\label{table:dataset_statistics}
\end{table}

\section{Node importance  interpretability score}

In addition to assessing the quality of clustering performed by LaPool and DiffPool, we attempt to measure the interpretability of their predictions. We consider a setting in which the goal is to predict the presence of either Epoxides, Thioepoxides or  Aziridines substructures in molecular graphs. These three fragments correspond to structural alerts that are often indicative of molecular toxicity. We attempt to identify the relational inductive bias used by each model during prediction. In our setting, we define the interpretability of a model as its ability to focus on nodes that are directly relevant to the structural alerts and leverage that information for its prediction. In other words, we expect the most important nodes for the model prediction to correspond to nodes that are part of the structural alerts. We measure the importance of each atom toward the model prediction using the Integrated gradient method. Briefly, we compute perturbation of node and edge attributes over a continuous spectrum, then integrate the gradient of each of the model loss with respect to both the perturbed adjacency matrix and node features. Similar to saliency maps, we then take the sum of the absolute integrated gradients over each node as an approximate attribution score for the nodes. Finally, we compute the interpretability score using the  Precision-Recall AUC between measured importance and ground truth which is defined by a binary mask of nodes that are part of the structural alerts. The PR-AUC allows us to assess the node importance separation capacity of each model while taking imbalance into account. We only focus on positive predictions for each model. As an alternative to the Integrated Gradient, we also measure the interpretability score using Guided BackPropagation (see Table~\ref{table:explain2})

\begin{table*}[htb]
\centering
\caption{Comparison of prediction explainability based on average PR-AUC over node attribution using various explainability framework.}
\label{table:explain2}
\footnotesize\begin{tabular}{lcccc}
\toprule
 & GIN & DiffPool & Graph U-net & LaPool  \\
\midrule
 Integrated Gradient &    0.876 & 0.799 & 0.879 & \textbf{0.906} \\
Guided BackPropagation &    0.819 & 0.739 & 0.835 & \textbf{0.857} \\
\bottomrule
\end{tabular}
\end{table*}

\section{Signal preservation through Laplacian maxima}\label{interpolate}

\begin{figure*}[ht]
    \centerline{\includegraphics[width=0.85\linewidth]{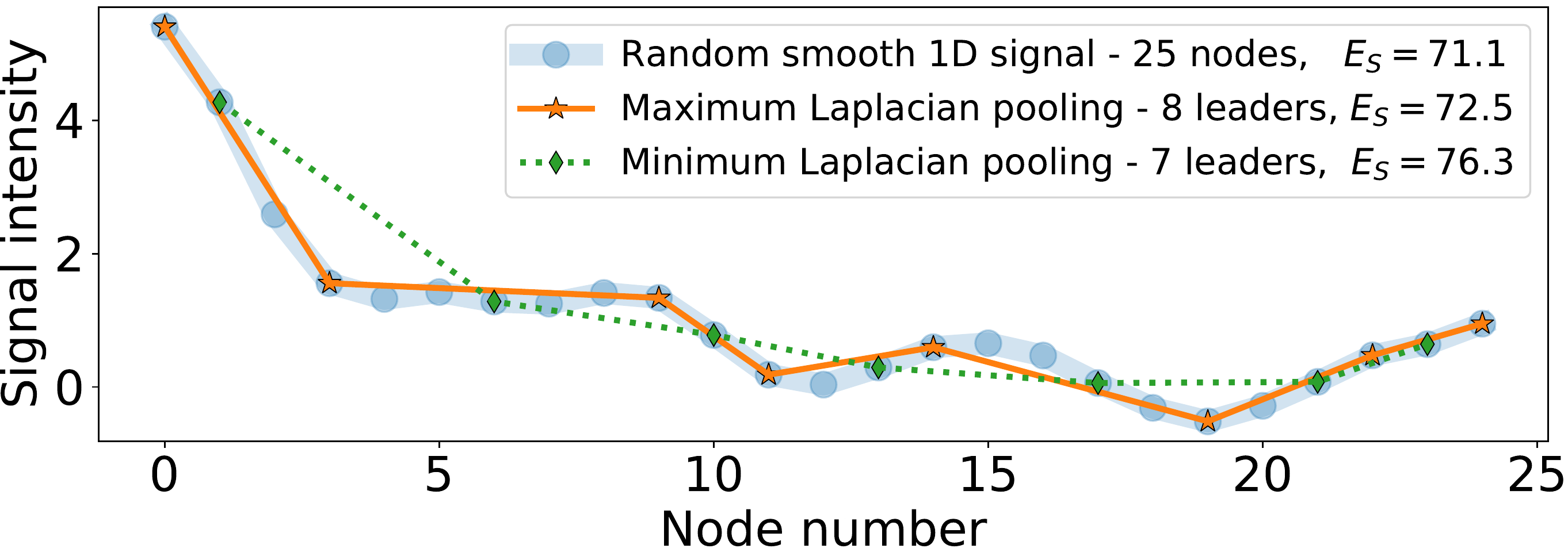}}
  \caption{Comparison of maximum/minimum Laplacian pooling for a random and smoothed signal on a 1D graph with 25 nodes. The graph energy $E_S$ is indicated.}    \label{fig:1D-signal}
\end{figure*}

We illustrate here on a 1-d signal $S$, how using the Laplacian maxima serves to retain the most prominent regions of the graph signal, after smoothing (Figure \ref{fig:1D-signal}). We measure the energy conservation after downsampling: $\delta_E(S) = E(S) -  E(S_{down})$ of the 1-d signal energy to highlight why selecting the Laplacian maxima allows reconstructing the signal with a low error when compared to the minimum Laplacian (which focuses on low frequencies). The energy $E_S$ of a discrete signal $y_i$ is defined in Equation (\ref{eq:E-graph}), and is similar to the energy of a wave in a physical system (without the constants). 

\begin{align}\label{eq:E-graph}
  E_S &= \sum_i{{|y_i|}^{2}}
\end{align}

To mimic the molecular graph signal at the pooling stage, the given signal is built from an 8-terms random Fourier series with added Gaussian noise, then smoothed with 2 consecutive neighbor average smoothing. For the pooling methods, a linear interpolation is used to cover the same signal space before computing $E_S$. As expected, the maxima Laplacian selection requires a fewer number of samples for signal reconstruction and  energy preservation. It also significantly outperforms minima selection.
\end{appendices}
\end{document}